\documentclass[double]{article}
\usepackage{times}
\usepackage{algorithm}
\usepackage{algorithmic}
\usepackage{wrapfig}
\usepackage{verbatim}
\usepackage{amsmath}
\usepackage{graphicx}
\usepackage{subcaption}
\usepackage[active]{srcltx}
\usepackage{color}
\usepackage{lineno}
\usepackage{dirtytalk}
\usepackage{amsthm}
\usepackage{authblk}
\usepackage{listings}
\usepackage{tikz}
\usepackage{longtable}
\usetikzlibrary{shapes.geometric, arrows}

\usepackage{epsfig,graphicx,amsfonts}
\usepackage{amsmath,amsthm,amssymb}
\usepackage{xcolor}

\usepackage{adjustbox}
\usepackage{multirow}
\usepackage{setspace}
\usepackage{hyperref}
\usepackage{comment}

\long\def\comment #1\commentend{}

\begin{document}

\title{\Large Predicting Postoperative Nausea And Vomiting Using Machine Learning: A Model Development and Validation Study}

\author{Maxim Glebov$^{1*x}$, Teddy Lazebnik$^{2,3*x}$,  Boris Orkin$^{4}$, Haim Berkenstadt$^{1,5}$, Svetlana Bunimovich-Mendrazitsky$^{2}$\\
\(^1\) Department of Anesthesiology, Sheba Medical Center, Ramat Gan, Israel\\
\(^2\) Department of Mathematics, Ariel University, Ariel, Israel\\
\(^3\) Department of Cancer Biology, Cancer Institute, University College London, London, UK\\
\(^4\) Digital Medicine and Technology, Holon Institute of Technology, Holon, Israel \\
\(^5\)  Faculty of Medicine, Tel-Aviv University, Israel\\
\(^*\) Corresponding author: hlebau@gmail.com, lazebnik.teddy@gmail.com \\
\(^x\) These authors contributed equally

}

\date{ }

\maketitle 

\begin{abstract}
\noindent
\textbf{Background:} Postoperative nausea and vomiting (PONV) is a frequently observed complication in patients undergoing surgery under general anesthesia. Moreover, it is a frequent cause of distress and dissatisfaction during the early postoperative period. The tools used for predicting PONV at present have not yielded satisfactory results. Therefore, prognostic tools for the prediction of early and delayed PONV were developed in this study with the aim of achieving satisfactory predictive performance. \\

\noindent
\textbf{Methods:} The retrospective data of adult patients admitted to the post-anesthesia care unit after undergoing surgical procedures under general anesthesia at the Sheba Medical Center, Israel, between September 1, 2018, and September 1, 2023, were used in this study. An ensemble model of machine learning algorithms trained on the data of 54848 patients was developed. The \textit{k}-fold cross-validation method was used followed by splitting the data to train and test sets that optimally preserve the sociodemographic features of the patients, such as age, sex, and smoking habits, using the Bee Colony algorithm. \\

\noindent
\textbf{Findings:} Among the 54848 patients, early and delayed PONV were observed in 2706 (4.93\%) and 8218 (14.98\%) patients, respectively. The proposed PONV prediction tools could correctly predict early and delayed PONV in 84.0\% and 77.3\% of cases, respectively, outperforming the second-best PONV prediction tool (Koivuranta score) by 13.4\% and 12.9\%, respectively. Feature importance analysis revealed that the performance of the proposed prediction tools aligned with previous clinical knowledge, indicating their utility. \\

\noindent
\textbf{Interpretation:} The machine learning-based tools developed in this study enabled improved PONV prediction, thereby facilitating personalized care and improved patient outcomes. \\

\noindent
\textbf{Funding:} This paper represents an independent research part-funded by Ariel University and the Holon Institute of Technology (grant number RA2300000519). \\ 

\noindent
\textbf{Keywords:} clinical machine learning; postoperative nausea prediction; personalized medicine. \\ 
\end{abstract}

\maketitle \thispagestyle{empty}
\pagestyle{myheadings} \markboth{Draft:  \today}{Draft:  \today}
\setcounter{page}{1}

\section{Introduction}
\label{sec:introduction}
Postoperative nausea and vomiting (PONV) is frequently observed in patients undergoing surgery under general anesthesia (GA) \cite{intro_1}. The risk of PONV has been reported to be 30\% and 80\% in the general surgical population and high-risk cohorts, respectively \cite{intro_14}. PONV can influence patient satisfaction with anesthesia and surgery, prolong the duration of stay in the post-anesthesia care unit (PACU), increase the incidence of unplanned admissions after outpatient surgery, and increase the costs associated with medical treatment \cite{haim3}. 

Previous studies have investigated the causes, prevalence, prevention, and treatment of PONV and developed evidence-based guidelines for the prevention and management of PONV \cite{intro_13}. The \textit{Apfel simplified risk score} \cite{intro_14} and the \textit{Koivuranta score} \cite{intro_15} have been proposed for the PONV risk assessment in the latest version of the guidelines \cite{intro_13}. However, despite their simplicity, these scores yield a predictive performance less than 70\% on average, which is unacceptable \cite{intro_16}. Therefore, a better tool is required for the prediction of PONV to facilitate an accurate assessment of patient risk and the formulation of evidence-based care for individual patients. 

Machine learning algorithms have been used increasingly to develop predictive models since the rise of artificial intelligence (AI) \cite{pick_tau_1}. These models have been show to outperform previous models based on classical statistics \cite{intro_nature}. Therefore, this study aimed to develop a model for predicting the risk of early (during PACU stay) and delayed (first 24 postoperative hours) PONV. Furthermore, the performance of the proposed model was compared with that of currently used prediction scores.

\section{Methods and Materials}
\label{sec:methods}
The protocol of this study was reviewed and approved by the Ethics Committee of the Sheba Medical Center, Israel (SMC 9646-22, January 25, 2023). This study was conducted in accordance with the principles of the Declaration of Helsinki of the World Medical Association.  The requirement for obtaining informed consent from the patients was waived by the Ethical Committee.
 
\subsection{Data Collection}
Data from our electronic patient records system, including biometric, medical, procedural, and physiological variables of patients anesthetized at our institution, were extracted and analyzed retrospectively. Patient data were anonymized and deidentified before being accessed and analyzed. All adult patients (age \(>\)18 years) admitted to the PACU who had undergone surgical procedures under GA, GA with neuraxial anesthesia (NA), GA with peripheral nerve block (PNB), or GA with NA and PNB at the Sheba Medical Center, Israel, between September 1, 2018, and September 1, 2023, were eligible for inclusion in this study. The exclusion criteria were as follows: patients who underwent surgery under local anesthesia, PNB, NA, and/or light sedation only; patients who underwent cardiac surgery or obstetric procedures, including cesarean section; patients with American Society of Anesthesiologists (ASA) physical status classification of grade 5; and patients who arrived intubated or required postoperative mechanical ventilation. Patients with perioperative medical records containing insufficient data were also excluded. PONV was defined as any documented event requiring the administration of a rescue antiemetic medication. Early PONV was defined as any documented event requiring the administration of a rescue antiemetic medication in the PACU, whereas delayed PONV was defined as an event requiring the administration of a rescue antiemetic medication during the first 24 postoperative hours.  

\subsection{Data analysis}
\begin{figure}[!ht]
    \centering
    \includegraphics[width=0.99\textwidth]{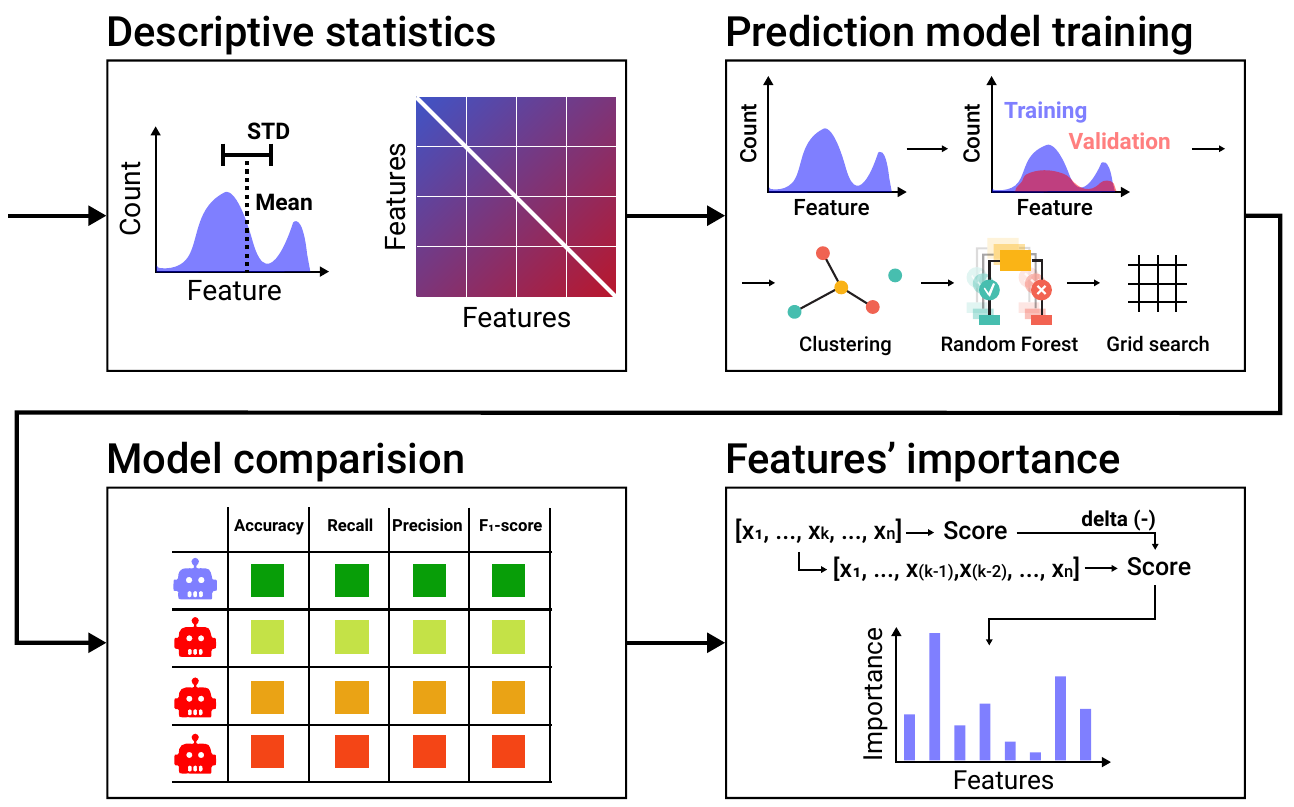}
    \caption{A schematic view of the workflow of the proposed framework.}
    \label{fig:scheme}
\end{figure}

The collected data underwent a three-step analysis. The statistical properties of the dataset were computed and analyzed initially. The data were subsequently divided into training and validation cohorts to facilitate the training and evaluation of the prediction tool, followed by the training of two machine learning-based algorithms for the prediction of early and delayed PONV. In addition, the performance of the obtained prediction tools was compared with that of the currently available PONV prediction scores. Lastly, the importance of each parameter of the prediction tool in learning the clinical reasoning revealed by these tools was evaluated based on the prediction tools obtained. All analyses were performed using the Python programming language (version 3.9). Fig. \ref{fig:scheme} provides a schematic of the workflow of the proposed framework.

\subsubsection{Cohort analysis}
The cohort data were statistically investigated initially by computing the mean, standard deviation, and median values of the continuous features and the distribution of categorical features. In addition, Pearson and Spearman correlation matrices from the cohort were computed to explore the linear and monotonic dependencies among the features and between the features and targets (i.e., early and delayed PONV). 

\subsubsection{Prediction tool training}
The study population was split into a training cohort (from which the proposed algorithm was derived) and a validation cohort (from which the prediction tools were applied and tested) to develop a machine-learning-based prediction tool. A popular cross-validation approach \cite{cross_validation}, which repeats the splitting of the study population into training and validation groups multiple times, was used to achieve a more statistically resilient evaluation of the performance of the prediction tool \cite{kfold_2}. In particular, the k-fold cross-validation method, which splits the data into \(k\) identical-sized and pairwise distinct cohorts, was used in this study. Each of these cohorts was used as the validation cohort once, and the remaining cohorts were used as training cohorts, resulting in \(k\) validations. The average performance of the prediction tool in these validation cohorts was computed to estimate the performance of the prediction tool.
The population was split into training and validation cohorts such that the age and sex distributions of both cohorts were statistically similar to ensure that the validation set follows a distribution similar to that of the training set, as required in clinical-related machine learning analysis \cite{split_important}. The age and sex distributions of each cohort should be similar to those of all other cohorts in the k-fold cross-validation method.
This condition is formalized as an optimization task such that a dataset, \(D\), is divided into \(k\) size-identical and pairwise distinct subsets. This minimizes the average distance between the distributions defined by the age and sex distributions of each cohort to those of any other cohort. Intuitively, this task is a private case of the nurse scheduling problem \cite{tech_2_2_1} (which is known to be NP-hard \cite{tech_2_2_2}). Based on \cite{tech_2_2_3}, a close-to-optimal solution was achieved using the Directed Bee Colony Optimization algorithm \cite{tech_2_2_4}. An ensemble of machine learning and feature selection algorithms were analyzed after each split in the training and validation cohorts to maximize the accuracy of the prediction tool. The k-fold cross-validation method was used in the training cohort to ensure that the obtained prediction tool is robust for the prevention of overfitting of the prediction tool on the training cohort and to improve data stability \cite{kfold_2}.
he divided training cohort was analyzed using a Tree-based Pipeline Optimization Tool (TPOT) \cite{tpot}, which is an automated machine learning tool that optimizes machine learning pipelines using genetic programming \cite{genetic_decision_making}. Furthermore, the hyperparameters of the model were tuned using the grid-search \cite{grid_search} method to improve its performance in terms of accuracy. Namely, multiple combinations of all hyperparameters of the prediction tool were sampled to determinate the hyperparameters' values that optimize the average accuracy of the k-fold cross-validation examination over the training cohort \cite{ml_4}. In addition, post-pruning methods were applied to the tree-based models to further improve the generalization and performance of these prediction tools \cite{teddy_sat}.

\subsubsection{Feature importance analysis}
The importance of the parameters was evaluated using the information gain method \cite{information_gain}. For each parameter used by the prediction tool, a feature was removed each time and the prediction tool was re-trained such that the average accuracy obtained from the k-fold cross-validation analysis was stored, resulting in an accuracy score for each removed parameter. A new parameter was introduced to the prediction tool subsequently, which was generated by sampling normally distributed noise with a mean of 0 and a standard deviation of 1. The decrease (or increase) compared with the accuracy of the prediction tool with all the parameters and without the “noise” parameter was computed for all these cases. All parameters with absolute differences smaller than the one obtained from the “noise” parameter case were set to zero. 
All values were normalized such that their sum was equal to one (i.e., \(L_1\) normalization) to obtain the importance of the parameters for each instance of the prediction tool. In addition, SHapley Additive exPlanations (SHAP) analysis was used to gain insight into the influence of various features on the obtained prediction tools \cite{clinical_shap}. The SHAP values can be used to explain the output of a machine-learning model by attributing the contribution of each individual feature to a particular prediction \cite{shap_value}. SHAP analysis originated in game theory and provides a method to estimate the contribution of features to the model's final prediction. The SHAP values quantify the extent to which each feature influences a prediction in the feature importance analysis. A positive SHAP value for a feature indicates that it contributes positively to prediction, whereas a negative value indicates that it has a negative impact.

\section{Results}
\label{sec:results}

\subsection{Descriptive statistics}
\label{sec:stats}
During the five-year period, 104,102 adult patients underwent surgery under anesthesia at our institution. The final cohort included 54,848 patients. 
Table \ref{table:descreptive} summarizes the descriptive statistics of the entire sample population. Continuous features are presented as count, range, mean value, and standard deviation values, whereas discrete features are presented as count and distribution. The Appendix summarizes the descriptive statistics of the early and delayed PONV subgroups.

\begin{longtable}{p{0.15\textwidth}p{0.3\textwidth}p{0.15\textwidth}p{0.15\textwidth}p{0.1\textwidth}p{0.1\textwidth}}
\hline \hline
\textbf{Symbol} & \textbf{Continuous variables} & \textbf{N} & \textbf{Range} & \textbf{Mean} & \textbf{SD} \\ \hline \hline
AGE & Age, years & 54845 & 18.00-102.56 & 47.81 & 19.16 \\
 HEIGHT & Height, cm & 50698 & 140.00-206.00 & 167.82 & 9.37\\
 BMI & Body mass index, kg*m-2 & 49653 & 13.40-69.30 & 26.35 & 5.66\\
 ANES\_DUR & Anesthesia duration, min & 49553 & 5.00-975.00 & 110.12 & 96.24 \\
 MORPH\_MGKG & Intraoperative morphine dose, mg/kg & 52227 & 0.000-0.566 & 0.061 & 0.053 \\
 FENT\_MCGKG & Intraoperative fentanyl dose, mcg/kg & 51408 & 0.000-36.364 & 3.102 & 1.795 \\
 MIDAZ\_MGKG & Intraoperative midazolam dose, mg/kg & 52410 & 0.000-0.280 & 0.017 & 0.015 \\
NEO\_MGKG & Intraoperative neostigmine dose, mg/kg & 53841 & 0.000-0.132 & 0.012 & 0.018 \\
CRIST\_MLKG & Intraoperative cristalloids, ml/kg & 51381 & 0.000-225.352 & 19.807 & 15.100 \\
URINE\_MLKG & Intraoperative urine output, ml/kg & 54429 & 0.000-85.085 & 0.986 & 3.150 \\
\hline
\textbf{Symbol} & \textbf{Categorical variables} & \textbf{Value} & \textbf{N} & \textbf{Percentage} &  \\ \hline
GENDER & Gender \\ 
& & female & 31429 & 57.30 \\
& & male & 23418 & 42.70 \\
SMOKE\_STAT & Smoking status \\
& & non smoker & 40111 & 78.61\\
& & smoker & 10912 & 21.39 \\
PROCEDURE & Procedure\\
& & laparoscopic appendectomy & 1975 & 3.61\\
& & in vitro fertilization & 1968 & 3.60\\
& & laparoscopic cholecystectomy & 1768 & 3.23\\
OPER\_DEPT & Operating department\\
& & general surgery & 15406 & 28.16\\
& & orthopedic surgery & 8690 & 15.89\\
& & gynecology & 7786 & 14.23\\
& & urology & 4444 & 8.12\\
& & hand surgery & 3226 & 5.90\\
& & ENT & 2920 & 5.34\\
& & neurosurgery & 2556 & 4.67\\
& & gynecology (IVF) & 2130 & 3.89\\
& & vascular surgery & 2012 & 3.68\\
& & gynecologic oncology & 1388 & 2.53\\
& & thoracic surgery & 1333 & 2.44\\
& & plastic surgery & 1143 & 2.09\\
& & maxillofascial surgery & 820 & 1.50\\
& & ophtalmology & 451 & 0.82\\
& & gastroenterology unit & 110 & 0.20\\
& & other & 288 & 0.53\\
ASA\_CLASS & ASA physical class \\
& & I & 9867 & 25.98 \\
& & II & 17100 & 45.03 \\
& & III & 9974 & 26.27 \\
& & IV & 1032 & 2.72 \\
APFEL\_SCORE & Simplified Apfel score\\
& & 0 & 6791 & 12.38\\
& & 1 & 18434 & 33.61\\
& & 2 & 24240 & 44.20\\
& & 3 & 5217 & 9.51\\
& & 4 & 165 & 0.30\\
KOIV\_SCORE & Koivuranta score\\
& & 0 & 3816 & 6.96\\
& & 1 & 12718 & 23.19\\
& & 2 & 23090 & 42.10\\
& & 3 & 11926 & 21.74\\
& & 4 & 3186 & 5.81\\
& & 5 & 111 & 0.20\\
GUID\_RISK & Number of risk factors (according to Fourth Consensus Guidelines for the Management of Postoperative Nausea and Vomiting)\\
& & 0 & 1962 & 3.58\\
& & 1 & 11040 & 20.13\\
& & 2 & 17911 & 32.66\\
& & 3 & 16957 & 30.92\\
& & 4 & 5459 & 9.95\\
& & 5 & 1457 & 2.66\\
& & 6 & 61 & 0.11\\
SURG\_URG & Surgery urgency\\
& & elective & 46154 & 84.10\\
& & urgent & 7955 & 14.50\\
OR\_LOC & OR location\\
& & main operating rooms & 35003 & 63.82\\
& & day surgery operating rooms & 7968 & 14.53\\
& & gynecological operating rooms & 11876 & 21.65\\
LAPAROSC & Laparoscopic approach & & 11459 & 20.89\\
NV\_MGMT\_24H & Antiemetic treatment 24h before surgery & & 4257 & 7.76\\
HX\_PONV & History of PONV & & 1101 & 2.01\\
MIGRAINE & History of migraine & & 572 & 1.04\\
NG\_TUBE & Nasogastric tube & & 15649 & 28.53\\
INDUCTION & Induction\\
& & intravenous & 48939 & 89.58\\
& & RSI & 4621 & 8.46\\
& & inhalational & 147 & 0.27\\
& & intravenous + inhalational & 926 & 1.69\\
AIRWAY\_MGMT & Airway management\\
& & intubation & 34389 & 72.10\\
& & LMA & 12757 & 26.75\\
& & mask & 509 & 1.07\\
& & tracheostomy & 43 & 0.09\\
NEURAX\_ANES & Neuraxial anesthesia & & 505 & 0.92\\
PERIPH\_NB & Peripheral nerve block & & 7291 & 13.29\\
INHALE\_ANES & Inhalational anesthetic & & 42171 & 76.89\\
NITROUS & Nitrous oxide & & 757 & 1.38\\ 
EPHEDRINE & Ephedrine & & 8533 & 15.56\\ 
SPINAL\_OPI & Spinal opioid & & 27 & 0.05\\
EPID\_OPI & Epidural opioid & & 225 & 0.41\\ 
MEPERIDINE & Intraoperative meperidine & & 2352 & 4.29\\ 
TRAMADOL & Intraoperative tramadol & & 1346 & 2.45\\ 
L\_ACT\_OPI & Intraoperative long acting opioid & & 40869 & 74.51\\ 
PARACET & Intraoperative paracetamol & & 40503 & 73.85\\ 
PRE\_PARACE & Intraoperative paracetamol before incision & &  15488 & 28.24\\ 
DIPYRONE & Intraoperative dipyrone & & 37856 & 69.02\\ 
NSAIDS & Intraoperative NSAIDs & & 2901 & 5.29\\ 
KETAMINE & Intraoperative ketamine & & 549 & 1.00\\ 
DEXMED & Intraoperative dexmedetomidine & & 52 & 0.09\\
SUGAMMA& Sugammadex & & 5140 & 9.37\\ 
PONV\_DRUGS & Number of PONV prophylactic drugs\\
& & 0 & 8878 & 16.19\\ 
& & 1 & 20559 & 37.48\\ 
& & 2 & 23749 & 43.30\\ 
& & 3 & 1661 & 3.03\\ 
ADH\_PROPH & Guidelines adherent prophylaxis & & 13308 & 24.26\\ 
LIB\_ANTIEM & Liberal antiemetic prophylaxis approach & & 2109 & 3.85\\ 
DEXA & dexamethasone & & 26899 & 49.04\\ 
5HT3\_ANT & 5-HT3 antagonist & & 38549 & 70.28\\
METOCLO & metoclopramide & & 7592 & 13.84\\ 
POSTOPI\_PACU & Postoperative long acting opioid & & 10584 & 19.30\\
NONOPI\_PACU & Postoperative non opioid analgesic & & 4330 & 7.89\\ 
PAIN\_MOD & Pain above moderate in PACU & & 9378 & 18.08\\
PONV\_PACU & early PONV (in PACU) & & 2706 & 4.93\\ 
PONV\_24H & delayed PONV (first 24h) & & 8218 & 14.98\\
PONV\_24H\_DPT & Surgical departments where delayed PONV manifested\\
& & general surgery & 4657 & 56.67\\
& & gynecology & 805 & 9.80\\
& & orthopedic surgery& 584 & 7.11\\
& & gyn onco & 380 & 4.62\\
& & thoracic surgery & 309 & 3.76\\
& & urology & 295 & 3.59\\
& & ent & 251 & 3.05\\
& & gyn day surgery IVF & 246 & 2.99\\
& & plastic surgery & 226 & 2.75\\
& & neurosurgery & 219 & 2.66\\
& & hand surgery & 132 & 1.61\\
& & other & 50 & 0.61\\
& & vascular surgery & 48 & 0.58\\ 
& & ophtalmology & 15 & 0.18\\

\hline \hline

\caption{A summary of the descriptive statistics of the dataset, showing the features utilized by the prediction tools. Continuous features are presented as count, range, mean, and standard deviation values. Discrete features are presented as distributions. The full table, as well as subsets of early PONV, delayed PONV, and no PONV, is provided in Supplementary Materials.}
\label{table:descreptive}
\end{longtable}

\begin{figure}
    \centering
    \includegraphics[width=0.99\linewidth]{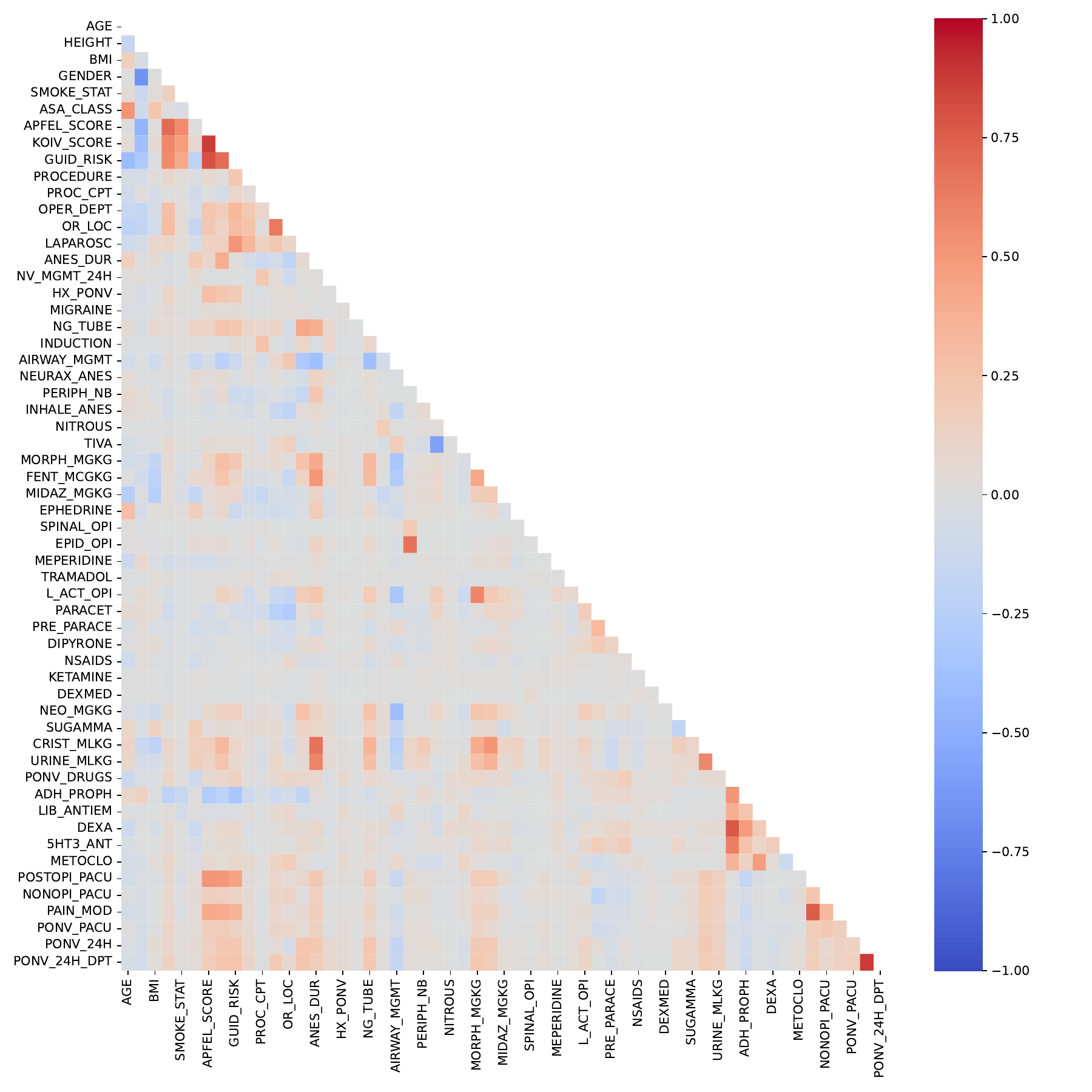}
    \caption{Pearson correlations between the dataset's features. }
    \label{fig:pearson}
\end{figure}

Fig. \ref{fig:pearson} shows the Pearson correlation matrix between the features of the prediction tool and themselves, including the target features. Most of the features were not correlated with each other, as the absolute values are close to zero. The Spearman correlation matrix provided in the Appendix indicates that both are similar. This finding indicates that the pairwise relationships between the features are linear and monotonic. Moreover, this finding shows that the task is complex and requires nonlinear models as a linear model would not be able to find meaningful relationships between the features \cite{teddy_black_economy}. A machine-learning approach that can capture complex and nonlinear datasets, especially in the clinical domain, was used in this study \cite{ml_3,ml_6}.  

\subsection{The perfomance of the proposed prediction tool and comparison with previous tools}
\label{sec:prediction_model}
\begin{figure}
\centering
\begin{subfigure}{.49\linewidth}
    \centering
    \includegraphics[width=0.99\linewidth]{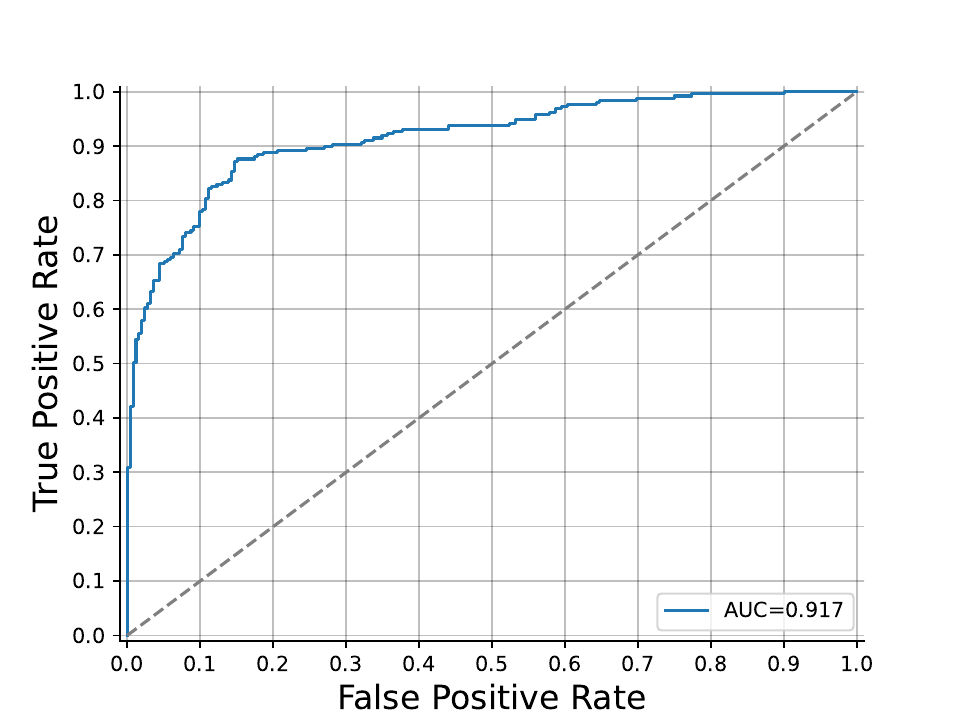}
    \caption{Early PONV}
    \label{fig:roc_pacu}
\end{subfigure}
\begin{subfigure}{.49\linewidth}
    \centering
    \includegraphics[width=0.99\linewidth]{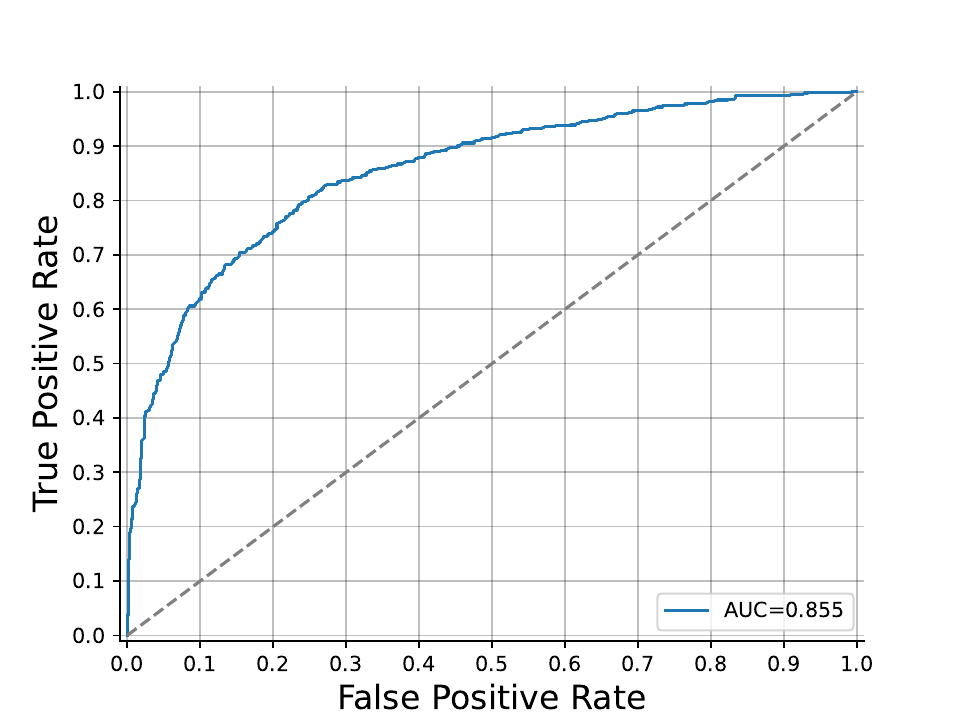}
    \caption{Delayed PONV}
    \label{fig:rpc_ponv}
\end{subfigure}
\caption{The receiver operating characteristic (ROC) curve of the obtained prediction tools for the early and delayed PONV tasks.}
\label{fig:roc}
\end{figure}

Two prediction tools were created: one to predict \say{early PONV} and one to predict \say{delayed PONV}. The receiver operating characteristic (ROC) curve and the area under the ROC curve (AUC) were calculated to evaluate the prediction tools, as presented in Fig. \ref{fig:roc}. The AUC scores were 0.917 and 0.855 for predicting early and delayed PONV, respectively. The accuracy, recall, precision, and \(F_1\)-score metrics for the proposed prediction tools were determined subsequently, following a \textit{k}-fold cross-validation method (\(k=5\)). In addition, in order to obtain a relative comparison to the currently used scores to evaluate the risk of early and delayed PONV, we also computed these metrics for the Apfel and Koivuranta scores.
The inclusion of the other model in this comparison was for the purpose of completeness, even though it is guidelines for the use of prophylactic antiemetics rather than genuine risk score to estimate the risk of PONV \cite{intro_13}. Table \ref{table:comparision} summarizes the results of this analysis. The proposed prediction tools outperformed all three scores for both tasks. The proposed prediction tool achieved an accuracy of 0.840 for predicting early PONV, whereas the other prediction scores achieved accuracies of 0.644, 0.706, and 0.536. Similarly, the proposed prediction tool achieved an accuracy of 0.773 for predicting delayed PONV, whereas the other prediction scores achieved accuracies of 0.570, 0.644, and 0.585. A one-sided ANOVA test performed for each task revealed that the proposed prediction tool showed statistically significant improvement in terms of accuracy (\(p < 0.001\) and \(p < 0.001\), respectively). 

\begin{table}[!ht]
\centering
\begin{tabular}{llllll} \hline \hline
\textbf{Task} & \textbf{Prediction tool} & \textbf{Accuracy} & \textbf{Recall} & \textbf{Precision} & \textbf{\(\boldsymbol{F_1}\)}\textbf{-score} \\ \hline \hline
\multirow{4}{*}{Early PONV} & Our & \textbf{0.840} &  \textbf{0.780} & \textbf{0.890} &  \textbf{0.831} \\
 & Apfel score & 0.644 & 0.417 & 0.777 & 0.543 \\
 & Koivuranta score  & 0.706 & 0.660 & 0.767 & 0.710 \\
 & Guidelines risk factors &  0.536 & 0.413 & 0.557 & 0.475 \\ \hline
\multirow{4}{*}{Delayed PONV} & Our &  \textbf{0.773} & \textbf{0.731} & \textbf{0.803} &  \textbf{0.765}  \\
 & Apfel score & 0.570 & 0.237 & 0.734 & 0.358 \\
 & Koivuranta score & 0.644 & 0.541 & 0.689 & 0.606 \\
 & Guidelines risk factors &  0.585 & 0.654 & 0.580 & 0.615 \\ \hline \hline
\end{tabular}
\caption{Comparison of the prediction performance of the proposed prediction tool with the simplified Apfel score, Koivuranta score, and risk factors according to the Fourth Consensus Guidelines for the Management of PONV. The results are shown as the mean of the \textit{k}-fold cross-validation analysis (\(k=5\)). The best prediction tool for each metric is highlighted in bold font.}
\label{table:comparision}
\end{table}

\subsection{Feature importance}
\begin{figure}
\centering

\begin{subfigure}{.49\linewidth}
    \centering
    \includegraphics[width=0.99\linewidth]{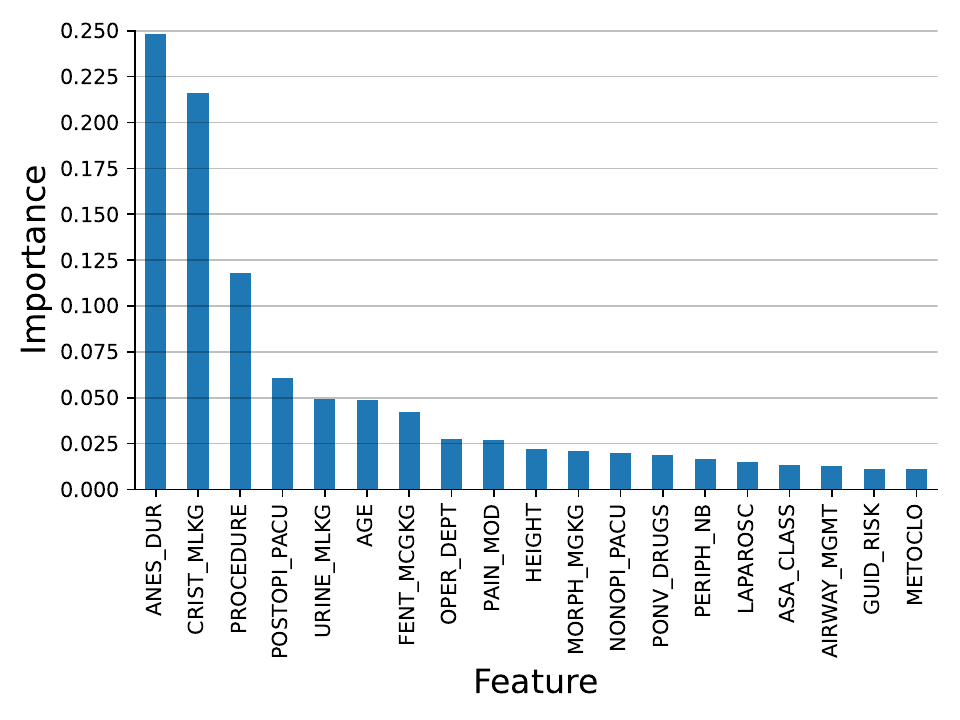}
    \caption{Early PONV.}
    \label{fig:fi_pacu}
\end{subfigure} 
\begin{subfigure}{.49\linewidth}
    \centering
    \includegraphics[width=0.99\linewidth]{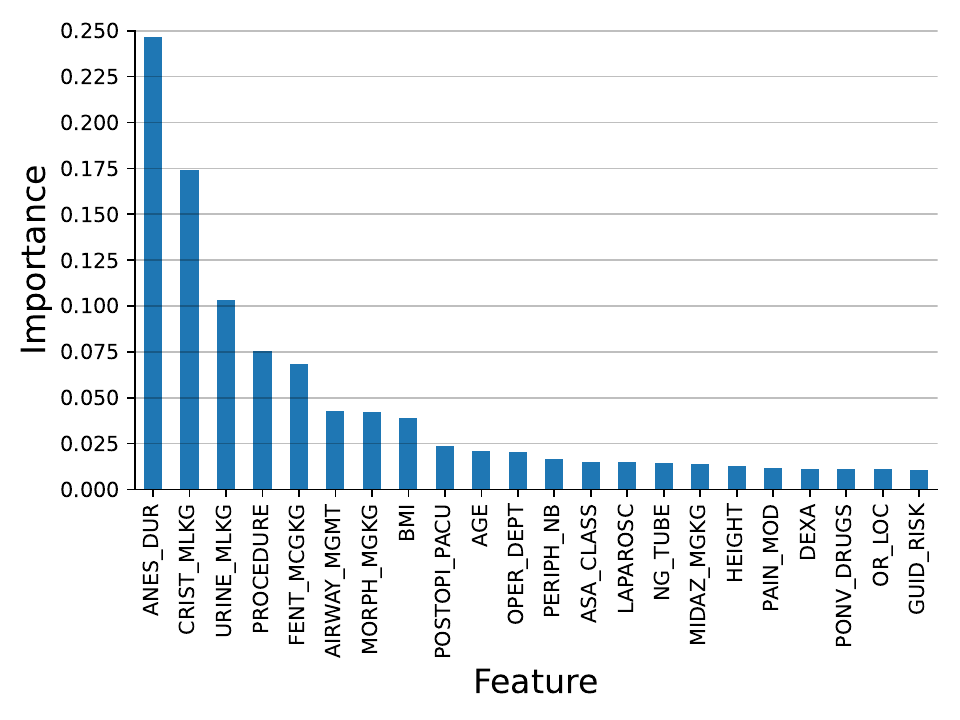}
    \caption{Delayed PONV.}
    \label{fig:fi_ponv}
\end{subfigure}
\caption{A feature importance analysis of the early and delayed PONV prediction tools.}
\label{fig:feature_importance}
\end{figure}

The information gain from each feature was computed for both tasks to determine its significance as a prediction tool. Fig. \ref{fig:feature_importance} presents the results of the analysis. Features with importance scores lower than those attributed to random noise were excluded for each prediction tool. The variables are arranged from left to right, and their order is determined based on their respective weights for the PONV effect. 
Fig. \ref{fig:fi_pacu} shows that the duration of anesthesia, volume of intraoperative crystalloids, type of procedure, administration of postoperative long-acting opioids, and intraoperative urine output were the five factors contributing the most to the incidence of early PONV. Fig. \ref{fig:fi_ponv} revealed that similar outcomes were significant factors for the incidence of delayed PONV.

\begin{figure}
\centering

\begin{subfigure}{.99\linewidth}
    \centering
    \includegraphics[width=0.99\linewidth]{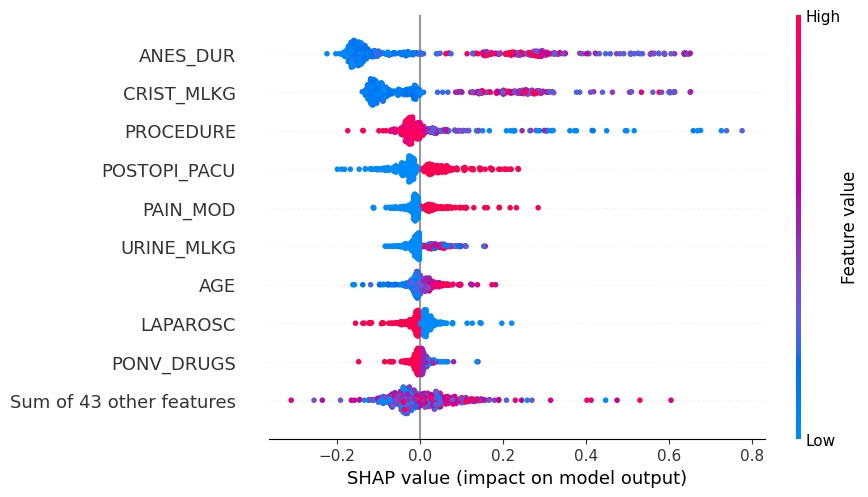}
    \caption{Early PONV}
    \label{fig:shap_pacu}
\end{subfigure}

\begin{subfigure}{.99\linewidth}
    \centering
    \includegraphics[width=0.99\linewidth]{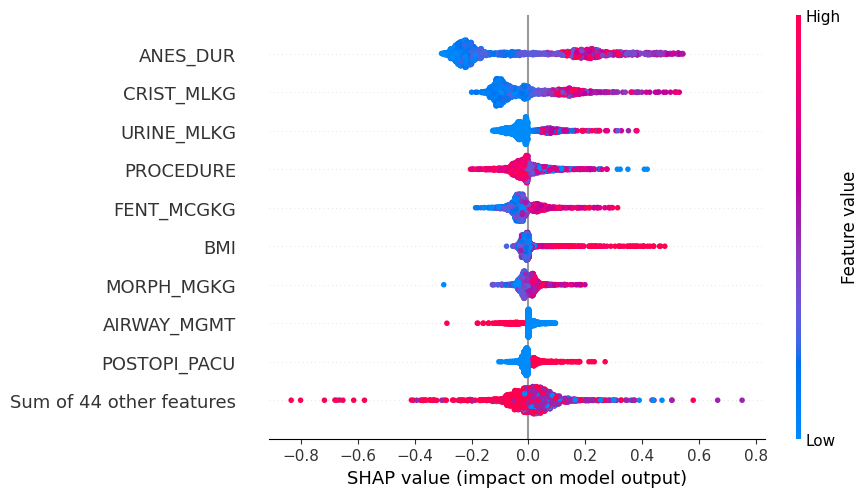}
    \caption{Delayed PONV}
    \label{fig:shap_ponv}
\end{subfigure}
\caption{A SHAP function value of each feature for the early and delayed PONV prediction tools.}
\label{fig:shap}
\end{figure}

The SHAP values \cite{clinical_shap} for both prediction tools were computed to obtain a better clinical understanding of the contribution of the variables to the model. Fig. \ref{fig:shap} shows the SHAP values of the top 10 combination variables. The color ranges from blue to red, indicating low to high values, and the y-axis indicates an increase or decrease in the probability of the incidence of early or delayed PONV according to each prediction tool. The early PONV prediction tool shown in Fig. \ref{fig:shap_ponv}. A longer duration of anesthesia, high volume of intraoperative crystalloids, specific types of procedures, administration of postoperative long-acting opioids, and pain in the PACU (above moderate level) showed associations with an elevated risk of early PONV. Fig. \ref{fig:shap_pacu} shows that a longer duration of anesthesia, high volume of intraoperative crystalloids, increased intraoperative urine output, specific procedure types, and cumulative dose of intraoperative fentanyl were associated with a higher risk of delayed PONV, as indicated by the mixture of blue and red dots.\\

\section{Discussion}
\label{sec:discussion}
The primary objective of this study was to develop a predictive model capable of assessing and forecasting the probability of PONV. Novel machine learning-based prediction tools were developed and validated using a comprehensive dataset from a diverse surgical population, including the records of over 54,000 patients, to predict the risk of early and delayed PONV. The findings of the present study revealed the complexity of PONV prediction as the analysis revealed multidimensional, nonlinear, and monotonic correlations among most variables and the risk of PONV (Fig. \ref{fig:pearson} and Table \ref{table:descreptive}). This underscores the importance of using statistical methods other than the traditional methods to utilize machine learning techniques, enabling us to capture the intricate relationships that influence PONV. 

The proposed prediction tools demonstrated excellent discriminative performance, as evidenced by the high AUC values of 0·917 and 0·855 for predicting the incidence of early and delayed PONV, respectively. A comparative analysis with classical PONV prediction scores, such as the Apfel and Koivuranta scores, revealed that the proposed prediction tools significantly outperformed these traditional approaches (Table \ref{table:comparision}). The accuracy values of 0·840 and 0·773 for predicting the risk of early and delayed PONV, respectively, were particularly encouraging when compared with those of the second-best prediction tool, the Koivuranta score, which achieved values of 0·706 and 0·644, respectively. Thus, these metrics suggest that the proposed prediction tools can effectively identify patients at risk for developing PONV, thereby enabling early intervention and personalized care. 

Consistent with the findings of previous studies, the present study revealed that the duration of anesthesia, type of procedure, the use of long-acting opioids, and the use of prophylactic medications are important predictors of early PONV \cite{intro_13}. The prediction tool developed for predicting delayed PONV has a similar distribution of feature importance. These findings validate the proposed model. However, in contrast with the findings of previous studies describing the protective effect of crystalloid infusion on PONV \cite{haim1}, the present study revealed that the volume of intraoperative crystalloids was an important independent predictor of PONV. A recent meta-analysis reported the protective effect is limited to healthy patients (ASA physical status 1–2) undergoing procedures that are ambulatory or require a short length of stay \cite{haim2}. The reasons for the occurrence of PONV due to this variable remain unclear and require further investigation.

This study has some limitations. First, the use of retrospective data from a single medical center may limit the generalizability of the prediction tools across different clinical settings. The proposed tools must be validated in multiple clinical contexts to confirm the robustness and external validity of the prediction tools. Multicenter studies with larger and more diverse datasets must be conducted to validate the prediction tools across different clinical settings. External validation is also necessary to ensure that the performance of prediction tools across various populations is preserved. Second, the present study focused on a patient population in Israel. Numerous studies have shown that ethnicity is an independent risk factor for PONV and that its incidence varies among different ethnic groups. The influence of ethnicity on the incidence of PONV could be influenced by pharmacogenomic and cultural factors \cite{discussion2}. 
Consequently, the proposed predictive tools cannot be generalized effectively to patient populations with distinct ethnic or cultural backgrounds. Future studies should include a more diverse patient cohort to ensure the broad applicability of predictive tools. Lastly, it is essential to recognize that the proposed prediction tools do not account for potential shifts in clinical practice or patient populations that could affect the incidence of PONV over time. These tools should be revisited periodically as new data are collected and medical practices change \cite{concept_drift}.

This study opens several avenues for future research. First, future studies should focus on the real-world implementation and clinical integration of the proposed prediction tools. By deploying and testing such “assistive” decision support platform integrated into the local electronic health record system, this study aimed to provide real-time notifications to anesthesiologists regarding high PONV-risk patients, thereby enhancing preventive actions and potentially improving outcomes. 
Future studies should aim to refine and expand the predictive capabilities of the proposed tools. Incorporating more diverse and comprehensive datasets, including genetic and pharmacogenomic information, may enhance the accuracy of risk assessments. Hence, the present study represents a significant step forward in the prediction and assessment of PONV in patients undergoing surgery. These machine learning-based prediction tools exhibited strong discrimination ability, clinical interpretability, and superior performance compared with those of traditional scoring systems. These findings hold significant promise in clinical practice, as they enable individualized PONV risk assessment and early targeted interventions, thereby improving patient outcomes and satisfaction.

\section*{Declarations}
\subsection*{Funding}
We gratefully acknowledge the financial assistance received from grant \#RA2300000519. The founders of the study had no role in study design, data collection, data analysis, data interpretation, or writing of the manuscript.

\subsection*{Conflicts of interest/Competing interests}
None.

\subsection*{Code and Data availability}
The code and data that have been used in this study are publicly available in this study's GitHub repository: \url{https://github.com/teddy4445/ponv_prediction_tool}.

\subsection*{Author Contribution}
Maxin Glebov: Conceptualization, Methodology, Formal analysis, Investigation, Data curation, Writing - Original Draft, Writing - Review \& Editing. \\  
Teddy Lazebnik: Methodology, Software, Formal analysis, Investigation, Writing - Original Draft, Visualization, Project administration. \\
Boris Orkin: Funding acquisition. \\ 
Haim Berkenstadt: Funding acquisition, Writing - Review \& Editing. \\ 
Svetlana Bunimovich: Funding acquisition. \\ 
 
\bibliography{biblio}
\bibliographystyle{plain}

\section*{Appendix}
\begin{figure}[!ht]
    \centering
    \includegraphics[width=0.99\textwidth]{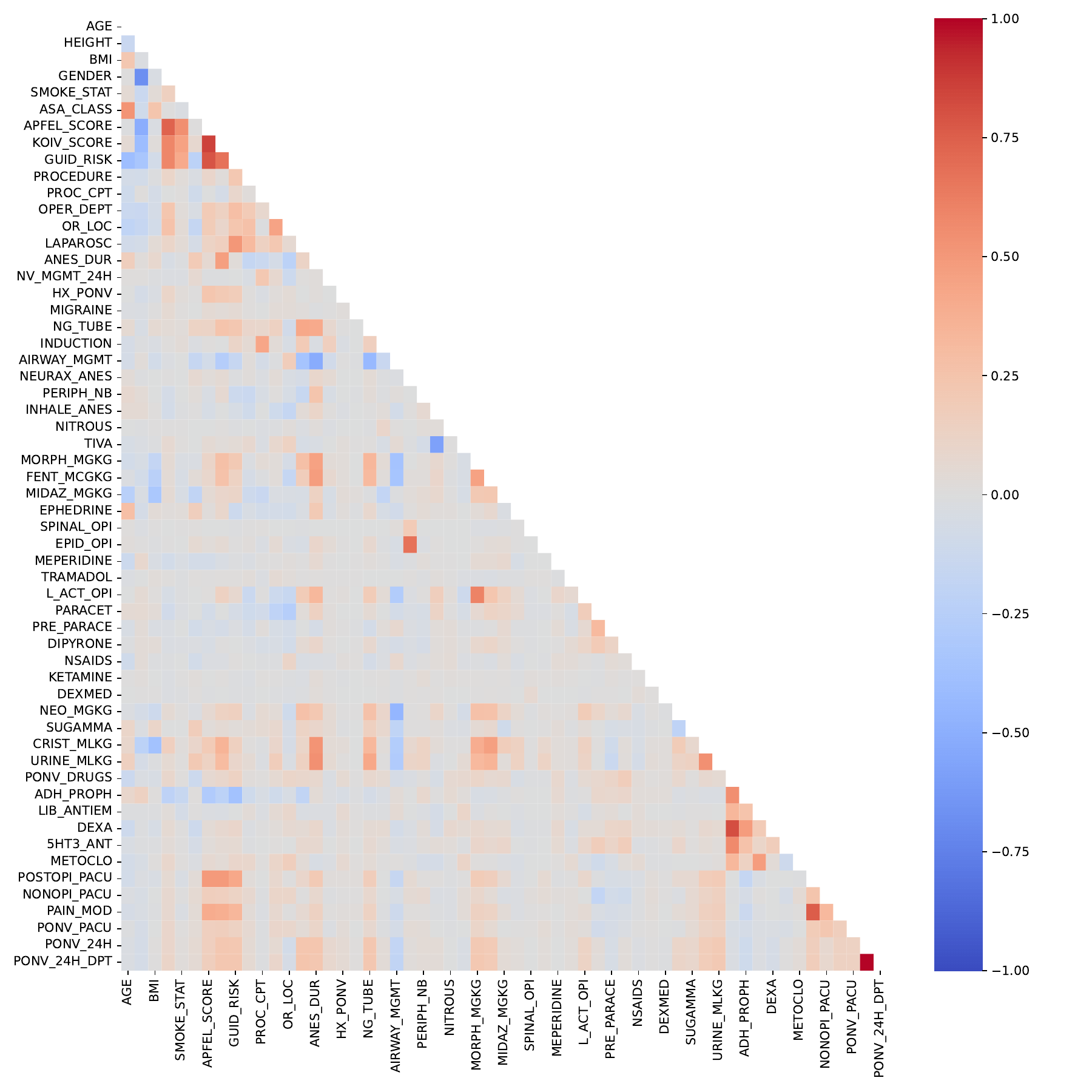}
    \caption{A Spearman correlation matrix between the features.}
    \label{fig:spearman}
\end{figure}

Fig. \ref{fig:spearman} shows the Spearman correlation matrix between the features of the model and themselves, including the target features. Most of the features showed no correlation with each other, as the absolute values were close to zero. The Spearman correlation matrix was very similar to the Pearson correlation matrix (Fig. \ref{fig:pearson}), indicating that the pairwise feature correlations were similarly linear and monotonic. 

\end{document}